\documentclass[sigconf]{acmart}

\AtBeginDocument{%
  \providecommand\BibTeX{{%
    \normalfont B\kern-0.5em{\scshape i\kern-0.25em b}\kern-0.8em\TeX}}}

\copyrightyear{2022}
\acmYear{2022}
\setcopyright{rightsretained}
\acmConference[CODS-COMAD 2022]{5th Joint International Conference on Data Science & Management of Data (9th ACM IKDD CODS and 27th COMAD)}{January 8--10, 2022}{Bangalore, India}
\acmBooktitle{5th Joint International Conference on Data Science \& Management of Data (9th ACM IKDD CODS and 27th COMAD) (CODS-COMAD 2022), January 8--10, 2022, Bangalore, India}\acmDOI{10.1145/3493700.3493742}
\acmISBN{978-1-4503-8582-4/22/01}

\newcommand{\RR}{\mathbb{R}}

\newcommand{\EE}{\mathbb{E}}

\begin{document}

\title[Local Linearity and Double Descent in Catastrophic Overfitting]{Local Linearity and Double Descent in Catastrophic Overfitting}

\author{Varun Sivashankar}
\authornote{Both authors contributed equally to this research. Both authors are current undergraduates.}
\email{varunsiva@ucla.edu}
\affiliation{%
\institution{University of California, Los Angeles}
\country{}
}

\author{Nikil Roashan Selvam}
\authornotemark[1]
\email{nikilrselvam@ucla.edu}
\affiliation{%
  \institution{University of California, Los Angeles}
\country{}
}


\begin{abstract}
  Catastrophic overfitting is a phenomenon observed during Adversarial Training (AT) with the Fast Gradient Sign Method (FGSM) where the test robustness steeply declines over just one epoch in the training stage. Prior work has attributed this loss in robustness to a sharp decrease in \textit{local linearity} of the neural network with respect to the input space, and has demonstrated that introducing a local linearity measure as a regularization term prevents catastrophic overfitting. Using a simple neural network architecture, we experimentally demonstrate that maintaining high local linearity might be \textit{sufficient} to prevent catastrophic overfitting but is not \textit{necessary.} Further, inspired by Parseval networks, we introduce a regularization term to AT with FGSM to make the weight matrices of the network orthogonal and study the connection between orthogonality of the network weights and local linearity. Lastly, we identify the \textit{double descent} phenomenon during the adversarial training process. 
Source code is available at \url{https://github.com/nikilrselvam/linearity-orthogonality-dd}.
\end{abstract}

\maketitle

\section{Introduction and Related Work}
Many deep learning models are vulnerable to \textit{adversarial attacks}. In particular, an image classifier can be tricked into predicting the wrong class for a given input by merely introducing small perturbations to the input\cite{inn,rof}. The most popular method to defend deep learning models from such adversarial attacks is Adversarial Training \cite{fgsm}, which can be formulated as a min-max optimization problem, where the inner maximization generates an adversarial example within the $\epsilon$-ball, and outer minimization optimizes the loss over the examples generated by the inner maximization. 

In practice, AT with \textit{Projected Gradient Descent} (PGD-10) provides the best adversarial robustness on several benchmark data sets. The optimization framework proposed in \cite{rof} can be formulated as:
\begin{equation}
\min_\theta \EE_{(x,y) \sim D} \left[\max_{\delta\in\Delta} \ell(x+\delta,y;\theta)\right]
\end{equation}
where $\Delta = \{\delta \in \RR^d : \|\delta\|_\infty \leq \epsilon\}$ is the $\ell_\infty$ $\epsilon$-ball in $\RR^d$.
Alternately, the \textit{Fast Gradient Sign Method} (FGSM)\cite{fgsm} has almost comparable performance with much less computing cost. Unlike PGD-10, FGSM solves the inner maximization problem in a single step\cite{fgsm}:
\begin{equation}
\delta_{\max} = \epsilon \text{sign}(\nabla_x \ell(x,y;\theta)) 
\end{equation}

However, recent work has shown that FGSM suffers from \textit{catastrophic overfitting} (CO), a phenomenon where during the training stage, often over just a couple of epochs, the test robustness declines sharply and often hits zero. This is accompanied by a sharp loss in \textit{local linearity} of the loss function with respect to the input space \cite{fat}. The authors of \cite{fat} were able to prevent CO and maintain high robustness by using the local linearity regularization term below, where $\cos(\cdot,\cdot)$ is the cosine of the angle between two vectors:
\begin{equation}
    \EE_{\eta \sim \mathcal{U}([-\epsilon,\epsilon]^d)}\left[1 - \cos(\nabla_x \ell(x,y;\theta), \nabla_x \ell(x+\eta,y;\theta)) \right]
\end{equation}

Another recent line of work \cite{par} has demonstrated that Parseval networks are very effective in preserving robustness. The weight matrices of such networks are approximately Parseval tight frames \cite{ort}, which are extensions of orthogonal matrices to non-square matrices. 
The authors show that every stochastic gradient update in the (non-adversarial) training regime preserves the orthogonality of the matrices by constraining the Lipschitz constants of all layers to be less than one which ensures high accuracy at low computational cost. 
Inspired by this, we use the Frobenius norm as regularization for the weights of the fully connected layers to investigate the effect of orthogonal weights on local linearity and robustness:
\begin{equation}
    \|W^T W - I\|_F^2 = \sum_{i=1}^{m}\sum_{j=1}^{n} |w_{ij}|^2 
\end{equation}

This motivates us to ask: \textit{Are local linearity and orthogonality of weight matrices correlated? Is either condition necessary/sufficient for preserving robustness and preventing catastrophic overfitting?}

Another phenomenon central to our work is double descent. It is characterized by three stages during training: a first descent in test loss followed by a sharp increase when the effective model complexity reaches a critical value, and then a second descent \cite{ddd}.
Double descent has been explored analytically in the case of linear regression \cite{ddl} and has been observed in deep neural networks \cite{ddd} but has not been well studied in the adversarial training regime.

\section{Main Contributions}
\begin{itemize}
    \item[1.] We empirically demonstrate that maintaining high local linearity may be \textit{sufficient} in preventing catastrophic overfitting but low local linearity does not necessarily cause it.
    \item[2.] Inspired by Parseval networks, we introduce an orthogonality regularization term to FGSM to study how orthogonality of network weights relates to local linearity.
    \item[3.] We identify the \textit{double descent} phenomenon during the adversarial training process. 
\end{itemize}

\newcommand\x{0.305}
\begin{figure}[]
\includegraphics[scale=\x]{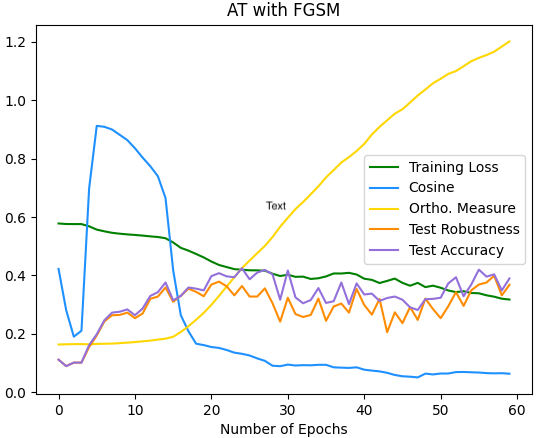}
\includegraphics[scale=\x]{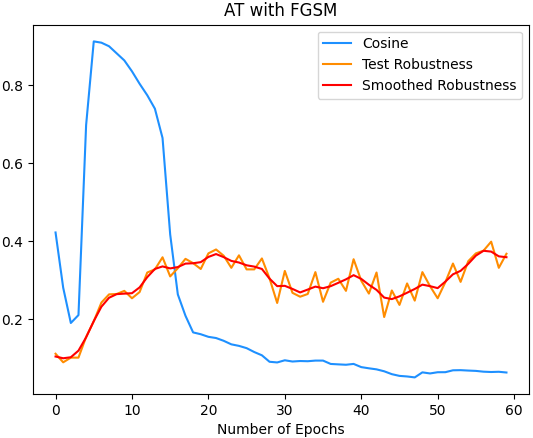}
\vspace{-10pt}
\caption{AT-FGSM, no regularization.}
\vspace{-10pt}
\label{fig:1} 
\end{figure}

\begin{figure}[]
\includegraphics[scale=\x]{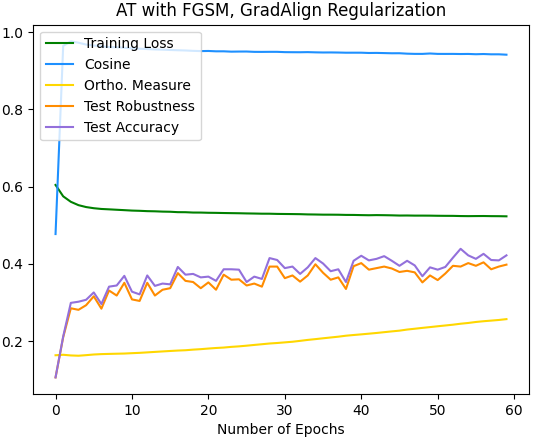}
\includegraphics[scale=\x]{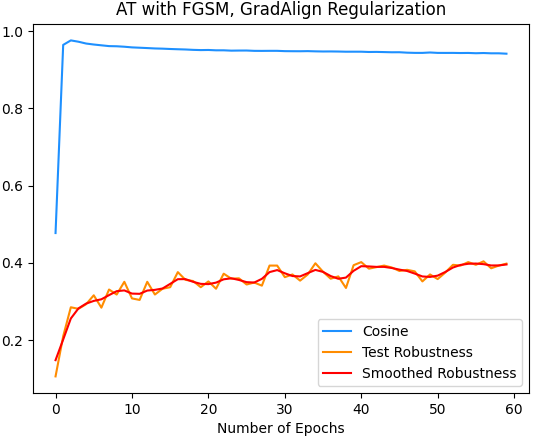}
\vspace{-10pt}
\caption{AT-FGSM, GradAlign regularization.}
\vspace{-10pt}
\label{fig:2} 
\end{figure}
\begin{figure}[]
\includegraphics[scale=\x]{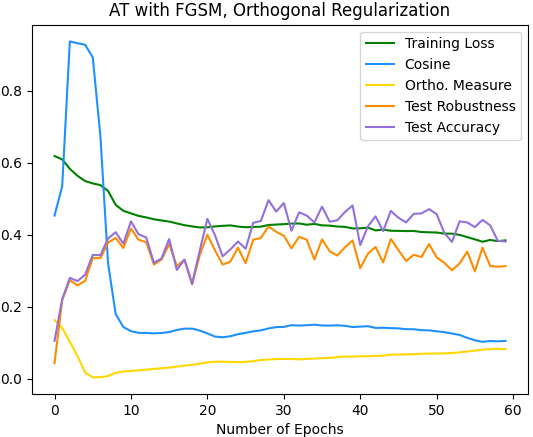}
\includegraphics[scale=\x]{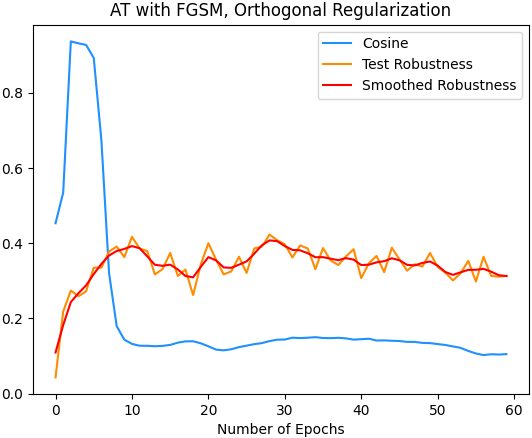}
\vspace{-10pt}
\caption{AT-FGSM, Orthogonal regularization.}
\vspace{-10pt}
\label{fig:3} 
\end{figure}
\begin{figure}[]
\includegraphics[scale=\x]{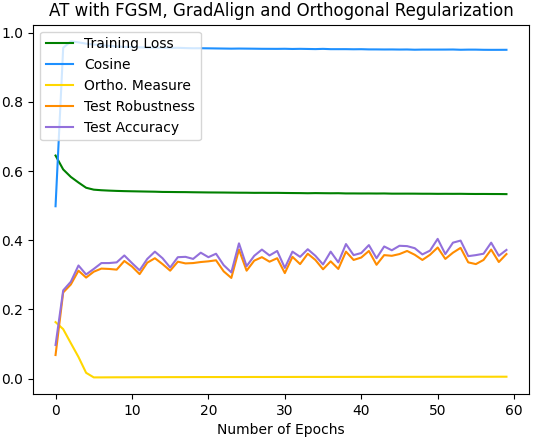}
\includegraphics[scale=\x]{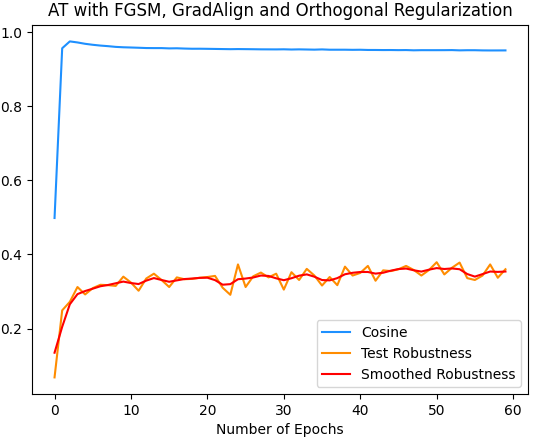}
\vspace{-10pt}
\caption{AT-FGSM, GradAlign \& Orthogonal regularization.}
\vspace{-10pt}
\label{fig:4} 
\end{figure}

\section{Experimental Results}


We conducted several experiments on CIFAR-10 \cite{cifar} using a 5-layer CNN:
First Convolutional Layer (3, 6, 5) with (2,2) Max Pooling, Second Convolutional Layer (6, 16, 5) with (2,2) Max Pooling, Three Fully Connected Layers (16 x 5 x 5, 120), (120, 84), (84, 10).

We opted for this architecture instead of Resnet-18 due to computational resource constraints. Of course, we were unable to replicate the performance of larger networks but we were able to clearly demonstrate qualitative trends that helped us address the questions raised in the previous section. We trained our model using AT with FGSM (with $\epsilon = \frac{16}{255}$) in 4 different regularization settings: GradAlign, Orthogonal, both, and none. The training data had 50,000 samples. The standard test set as well as the adversarial test set contained 1000 samples. The adversarial test set was constructed from the test set using a white box PGD-10 attack.
We draw the following inferences from our experiments.
\begin{itemize}
    \item[1.] Our experiments suggest that high local linearity is \textit{not necessary} for high robustness. This is seen in Fig. \ref{fig:1} and \ref{fig:3}, where the robustness continues to increase even after the local linearity has fallen considerably. Although robustness declines later in the training process, it eventually recovers even though local linearity is almost 0.
    \item[2.] In Fig. \ref{fig:2} and \ref{fig:4}, when high local linearity
is maintained, there is no decline in robustness. This validates that high local linearity is sufficient in preventing catastrophic overfitting.
    \item[3.] In Fig. \ref{fig:3}, when we try to keep orthogonality high, orthogonality and local linearity appear to move in the same direction. However, this does not prevent a fall in local linearity.
    \item[4.] Conversely, in Fig. \ref{fig:2}, we only control local linearity, but the orthogonality is significantly higher than in Fig. \ref{fig:1}. This suggests that high local linearity induces orthogonal weights.
    \item[5.] Importantly, in Fig. \ref{fig:1} and \ref{fig:3}, when local linearity is not explicitly controlled, we observe that the network \textit{undergoes overfitting but then subsequently recovers}. This happens from epoch 20 to 60 in Fig. \ref{fig:1} and from epoch 10 to 30 in Fig. \ref{fig:3}. This is characteristic of double descent. Additionally, \textit{the drop in robustness does not coincide with the drop in local linearity}, unlike what was observed in \cite{fat} with a much larger neural network.
\end{itemize}

\section{Conclusion and Future Work}
Contrary to previous work, we have empirically demonstrated that local linearity is not a necessary condition for preventing catastrophic overfitting but is likely sufficient, and may also induce orthogonal weights. While our work does not provide a new defense algorithm or technique, we believe it opens up plenty of interesting areas for future work:
\begin{itemize}
    \item Finding a new primary cause for catastrophic overfitting. 
    \item Formally investigating the double descent phenomenon in the adversarial training framework with statistical explanations for our empirical observations. 
    \item Exploring the potential use of Parseval tight frames in adversarial training. While \cite{par} studies the role of orthogonality in the standard training regime, it will be interesting to see the effect of restricting our weight space to those induced by tight frames in the adversarial training regime. 
\end{itemize}

\bibliographystyle{ACM-Reference-Format}

\end{document}